\documentclass[10pt,twocolumn,letterpaper]{article}

\usepackage{btas}
\usepackage{times}
\usepackage{epsfig}
\usepackage{graphicx}
\usepackage{amsmath}
\usepackage{amssymb}
\usepackage{xcolor}
\usepackage{multirow}
\usepackage{subcaption}

\DeclareMathOperator*{\argminB}{argmin}   
\btasfinalcopy 

\ifbtasfinal\fi
\begin{document}

\title{Demography-based Facial Retouching Detection using Subclass Supervised Sparse Autoencoder}

\author{Aparna Bharati$^1$, Mayank Vatsa$^2$, Richa Singh$^2$, Kevin W. Bowyer$^1$, Xin Tong$^1$\\
$^1$University of Notre Dame, $^2$IIIT-Delhi\\
{\tt\small $^1$\{abharati, kwb, xtong1\}@nd.edu},
{\tt\small $^2$\{mayank, rsingh\}@iiitd.ac.in}
}

\maketitle
\thispagestyle{empty}

\begin{abstract}

Digital retouching of face images is becoming more widespread due to the introduction of software packages that automate the task. Several researchers have introduced algorithms to detect whether a face image is original or retouched. However, previous work on this topic has not considered whether or how accuracy of retouching detection varies with the demography of face images. In this paper, we introduce a new Multi-Demographic Retouched Faces (MDRF) dataset, which contains images belonging to two genders, male and female, and three ethnicities, Indian, Chinese, and Caucasian. Further, retouched images are created using two different retouching software packages. The second major contribution of this research is a novel semi-supervised autoencoder incorporating ``subclass'' information to improve classification. The proposed approach outperforms existing state-of-the-art detection algorithms for the task of generalized retouching detection. Experiments conducted with multiple combinations of ethnicities show that accuracy of retouching detection can vary greatly based on the demographics of the training and testing images.
\end{abstract}

\section{Introduction}
\label{sec:intro}

\begin{figure}[t]
\begin{center}
\includegraphics[scale=.325]{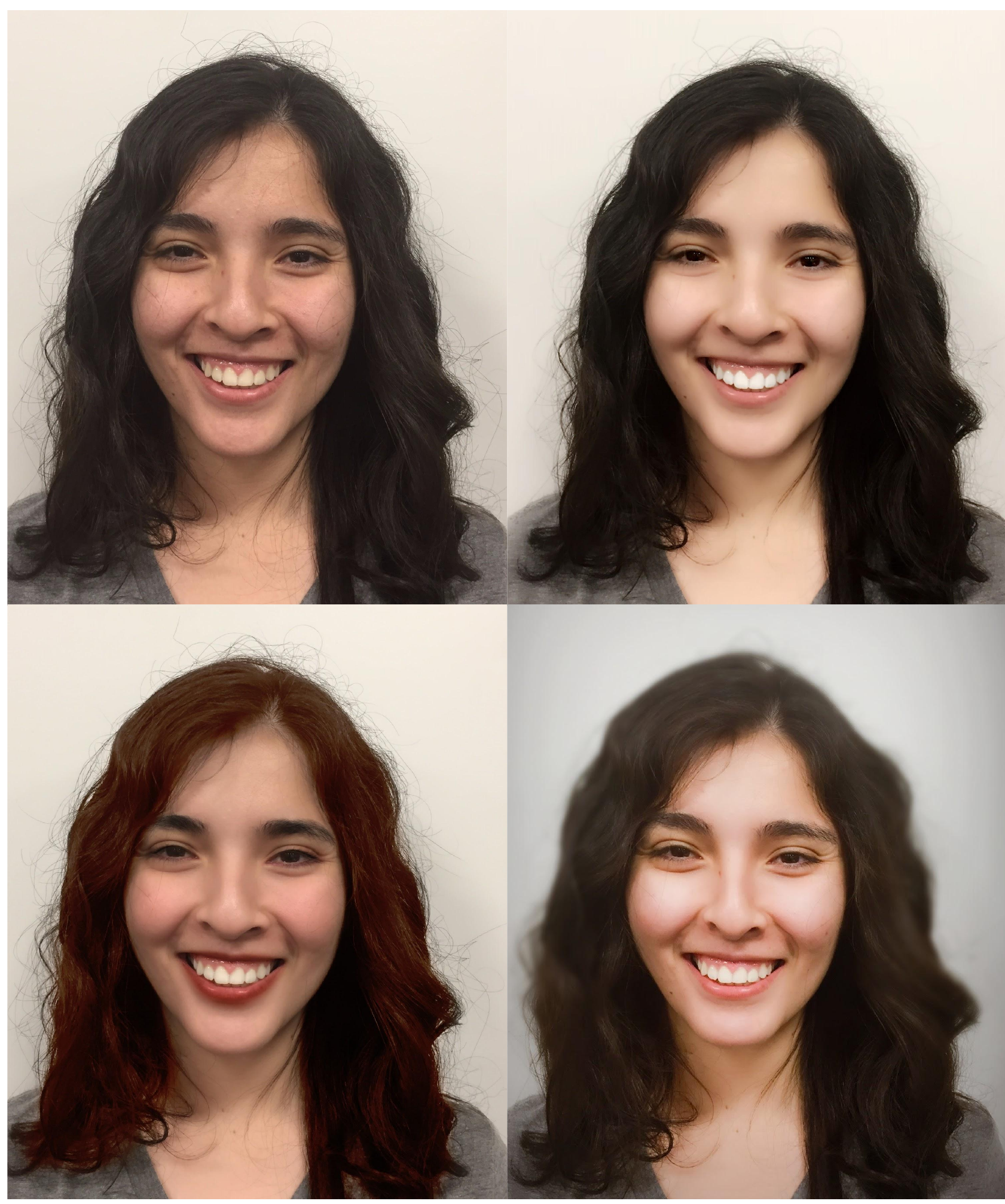}
\end{center}
\caption{Retouched samples of a face image using three different kinds of retouching tools. From top left to bottom right: original image, image retouched by \textit{Fotor} \cite{fotor}, \textit{PotraitPro Studio Max} \cite{potraitpro} and \textit{BeautyPlus} \cite{beautyplus}.}
\label{fig:retouch}
\end{figure}

People are increasingly using software such as Adobe Photoshop, Pixlr and Fotor in order to improve the appearance of images contributed to media-sharing websites.
Such processing when specifically applied to face images with the intent of improving attractiveness is what is considered digital facial retouching, as introduced in~\cite{bharati2016detecting}. Some effects of the process include smooth skin, change in color and geometry of facial features and improved image lighting. Figure~\ref{fig:retouch} presents few retouched versions of a face with some of these effects. 

The resultant image from retouching is a plausible variant of the input face which, depending on the extent of retouching, may or may not match well to the original face~\cite{bharati2016detecting}. When extensive retouching renders a face unrecognizable, this raises security concerns for face recognition systems~\cite{ferrara2013impact}. There are scenarios where the gallery images for the Automatic Border Control (ABC) systems are collected through web uploading or mailing of printed photographs. In these cases, the quality or originality of the photographs cannot be strictly assured or checked. This demands for an automatic system that can raise an alarm in case of unacceptable level of alterations in submitted photographs. 

Apart from the security concerns, there are also social concerns regarding the need for ideal appearance of an individual in the society~\cite{little2011facial}. A recent trend of retouched photos has been observed on online dating sites~\cite{dating}. People retouch their photos to look younger and slimmer. Israel has announced the \textit{`Photoshop Law'}~\cite{israel} which regulates advertisements and magazine covers with photoshopped images to include declarations. The legislation was inspired from the correlation between increased rates of anorexia and bulimia in the young female population and media portrayals of the ideal size and figure. After Israel, the issue received attention in other countries such as France, UK and USA~\cite{france, laws, selfesteem}. 
Therefore, an accurate automatic detector for retouching in images can help in better enforcement of these laws. 

\begin{figure}[t]
\centering
\includegraphics[scale=.35]{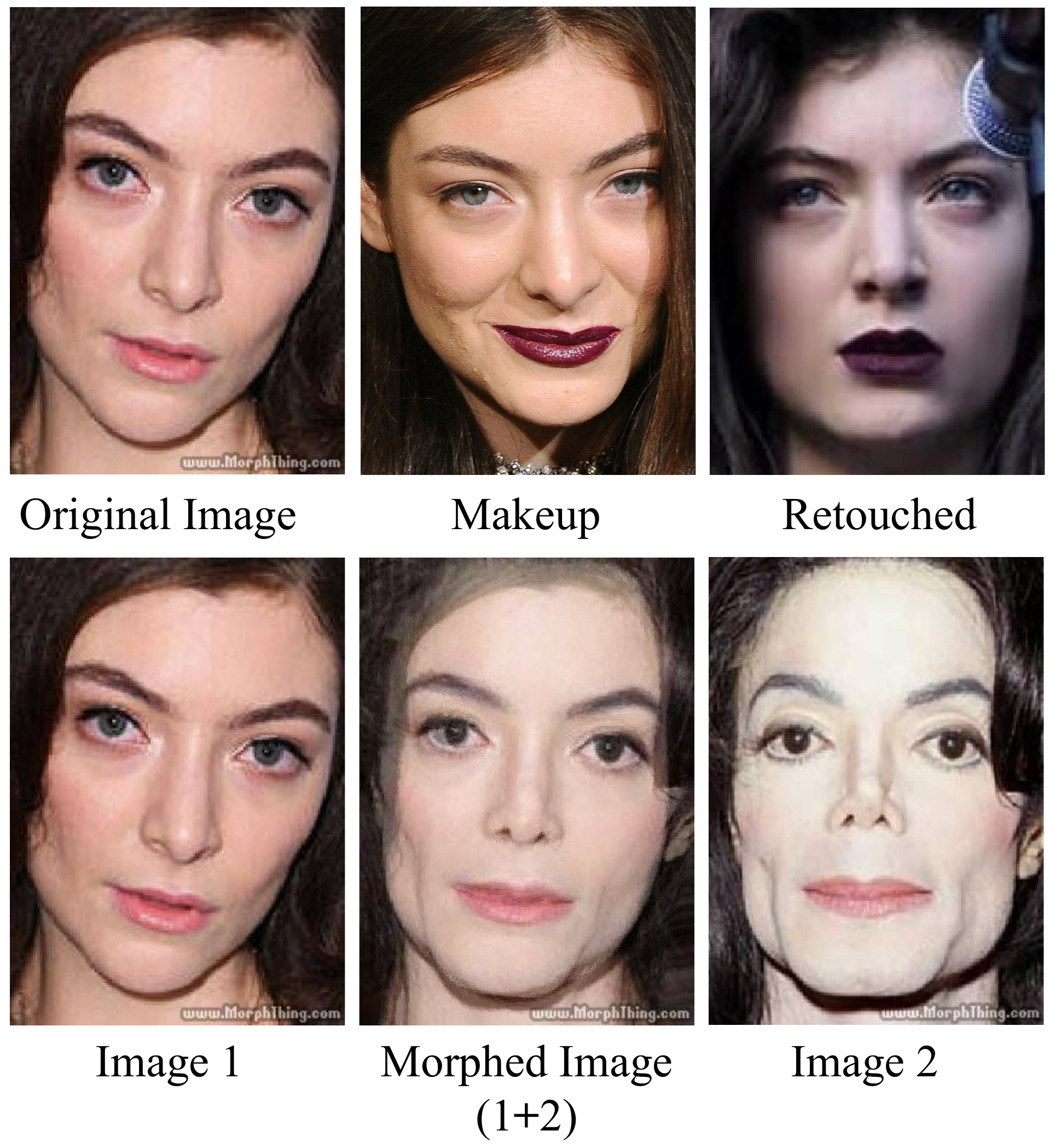}
\vspace{5pt}
  \caption{Examples of image with makeup and retouching (first row) and morphing (second row).}
\label{fig:spoof}
\end{figure}



Facial retouching if performed to impersonate another identity, like morphing, has the potential to affect face recognition systems~\cite{ferrara2013impact, raghavendra2016detecting}. Also, in a general scenario, the aim of retouching might not be to impersonate or spoof but to beautify the base face in a yet identifiable fashion~\cite{hanyinterview}. For this kind of retouching, the facial features might change to some extent but the face still remains closer to the original identity than others in the face space. With spoofing, the aim is to push the face image away from the real identity and closer to another identity. Therefore, the methods used for detecting morphing and other kinds of spoofing might not be sufficient to detect retouching.


Another important related issue with face images is makeup. Even though makeup and retouching share the purpose of beautification and can easily be confused with each other, they are different processes. Real makeup is applied to a face before image acquisition while retouching and virtual makeup are applied after the image is acquired. Though retouching may include digital application of makeup, it is not limited to it. Retouching can yield various effects which is not possible through virtual makeup. For example, it can change facial shape which a virtual makeup tool cannot. Also, the lighting and ambience of the face in the picture cannot be altered through virtual makeup but such an effect can be achieved through retouching. Figure~\ref{fig:spoof} shows an example illustrating the case of morphing of two celebrity faces and real makeup versus retouching for a face.

\subsection{Literature Review}
\label{subsec:related}

In the existing literature, there are efficient algorithms designed specifically to predict the presence of retouching for images such as magazine covers. Kee \etal~\cite{kee2011perceptual} propose a set of geometric and photometric features from face and body for retouching that correlates well with human perception. For profile pictures on social media and dating websites, the only information available can be obtained from face. The recent work by Bharati \etal~\cite{bharati2016detecting} is relevant to facial retouching detection. It uses a deep learning framework and outperforms several handcrafted features. In this paper, the authors introduced the first retouched faces database - ND-IIITD Retouched Faces database, with images from 325 subjects (mostly Caucasian) and retouched images with one tool. They proposed a Supervised Deep Boltzmann Machine (SDBM) to extract features and a $2\nu$ Support Vector Machine (SVM) for classifying the images into unaltered or retouched class. 

Other detection algorithms relevant to our work are algorithms for makeup detection and morphing detection. For automatic detection of makeup, Chen \etal~\cite{chen2013automatic} propose extracting color, shape and texture features of three pre-defined facial regions which are classified using SVM with RBF kernel and Adaboost. They perform experiments using YouTube Makeup (YMU) database for training and Makeup in the wild (MIW) database for testing.
Later, Kose \etal~\cite{kose2015facial} proposed a more accurate makeup detection algorithm on the same datasets using texture and shape features. 

A recent solution to detecting morphing for faces has been proposed by Raghavendra \etal~\cite{raghavendra2016detecting}. The proposed algorithm uses Binarized Statistical Image Features (BSIF) and linear SVM. They report results on a dataset of 450 morphed face images from 110 individuals. These detectors have state-of-the-art performances in detecting makeup and morphing.

\subsection{Research Contributions}
\label{subsec:contributions}
 

Coetzee \etal~\cite{coetzee2014cross} and Rhodes \cite{rhodes2006evolutionary} found that the race of an individual affects their perception of beauty or facial attractiveness. They also suggest that the perception varies between males and females. These might lead to different desired retouching effects for groups of individuals, implying different subclass distributions within both original and retouched classes.

Previous research on retouching detection algorithms has not focused on detecting retouching in various demographics. Also,
most of them only consider retouching effects provided by one tool which doesn't test the algorithm for tool biases. In this paper, we improve upon the existing literature by making the following key contributions:


\begin{itemize}

\item 
We introduce a Multi-Demographic Retouched Faces (MDRF) database which consists of retouched images from two retouching tools and three ethnicity variations.

\item A novel subclass-based supervised sparse autoencoder ($S^3A$) formulation is proposed to impart specialized sparse encoding to samples. We claim that this helps generate subclass specific representations. A comparison of the proposed method with the existing state-of-the-art \cite{bharati2016detecting} retouching detection algorithm is provided for three different ethnicities.

\item Through combined and cross-ethnicity performance analysis, we highlight the effects of different ethnicity, gender and tools on algorithms used for retouching detection.

 

\end{itemize}

\begin{figure*}[t]
\centering
\includegraphics[scale=0.30]{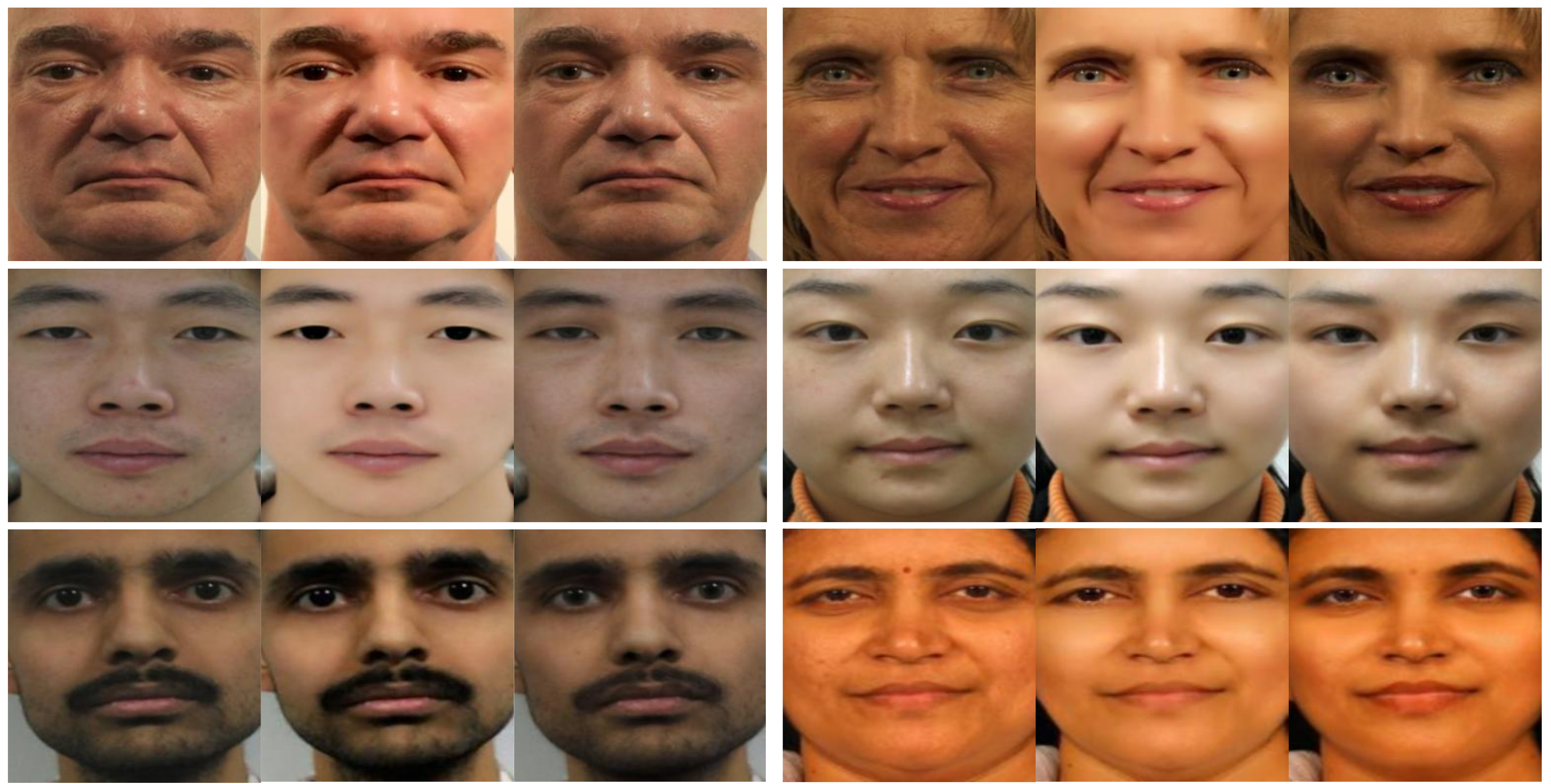}
\vspace{5pt}
   \caption{Sample images from the MDRF database. The first row consists images of Caucasian subjects, second row consists images of Chinese subjects and the third row consists of Indian subjects. For each of the six subjects, the first image is an original image, the second image is a retouched version of the first image using \textit{BeautyPlus}, and the third image is a retouched version of the first image using \textit{PotraitPro Studio Max}. The male images are on the left half and female images are on the right.}
\label{fig:samples}
\end{figure*}

\section{Multi-Demographic Retouched Faces Dataset}
\label{sec:dataset}

Datasets used in previous research on retouching detection do not well represent different races and generally contain retouched images from a single software package. Considering these variants for retouched images is important for a real-world scenario where the image in question can belong to any ethnicity and can have specialized retouching effects pertaining to the preferences of the individual or group. In order to analyze the effects of retouching on faces from different demographics and test performance of retouching detection algorithms, we created a Multi-Demographic Retouched Faces (MDRF) database. The database has a total of 3600 images, where 1200 belong to original class and 2400 belong to retouched class; 1200 belong to each of the Chinese, Indian and Caucasian classes; and 1800 each to the male and female classes. Sample images from the database are presented in Figure~\ref{fig:samples}. Further details about the tools and their retouching effects are provided next. 

\subsection{Retouching Tools}
\label{subsec:creation}
In order to capture the variance of retouching effects provided by various tools in the dataset, the retouched images were created using two tools. Both the tools have a different set of features or presets which lead to different effects. 

\textit{\textbf{BeautyPlus}} is a widely used (900 million users)~\cite{beautyplus} mobile phone application that helps users retouch pictures. 
It provides an array of features related to retouching such as skin smoothing, skin tone enhancement, acne removal, and face slimming~\cite{beautyplus}. For each original image, a set of effects to apply and the degree of application was randomly selected. This gives the liberty to the user to customize the effect for each image but it acts as a limitation since processing each image by hand is a tedious task and hinders large-scale generation of retouched images. 

\textit{\textbf{Potraitpro Studio Max}}~\cite{potraitpro} is a paid service available as a stand-alone desktop application as well as plug-in for Adobe Photoshop. The tool detects the specific face features such as - eyebrows, eye corners, lips and nose centers and uses this information to provide different changes to each part. There are built-in presets to be used as filters on specific parts - eyes, lips, nose, or the entire face. These presets can change color of the eyes and lips, brighten the eyes, teeth and skin, sculpt (change the shape) of the jaw, forehead or the full face. The tool allows the creation of new presets using a mix and match of the existing ones. Once the effect has been determined, it can be applied to a set of faces using the batch processing mode. This helps us automatically create more than one retouched image for multiple individuals.

\subsection{Distribution of Data}
The images in the MDRF database belong to two categories -  original and retouched, as defined below.

\begin{itemize}
    \item \textbf{Original} - The database includes two frontal images for each of the 600 subjects from same session as original images. The original images have been acquired under controlled conditions and are known to have not been retouched. The database has balanced distribution of images for two genders (100 males and 100 females subjects for each ethnicity) and three ethnicities (200 subjects each).
    
    The Caucasian subjects were selected from the Collection B of Notre Dame database~\cite{flynn2003assessment}. The original images for Chinese subjects were sampled from the visible spectrum images provided by the CASIA-NIR-vs-VIS (version 2) dataset~\cite{li2013casia}. The images for Indian subjects were selected from the CSCRV database \cite{singh2016cross}. 
    
    \item \textbf{Retouched} - The second image of each subject is used to create retouched images. Two tools were used to retouch images of all subjects - \textit{BeautyPlus} and \textit{Potraitpro Studio Max} (v12). For female subjects, we also created images with virtual makeup applied using an application similar to BeautyPlus, called MakeupPlus. For male subjects, two retouched images (with different combination of retouching options) were obtained from each of the tools while for female subjects, 2 images were obtained from BeautyPlus/MakeupPlus and 2 from Potrait Pro. This resulted in 2400 retouched images in total (4 images per subject).         
\end{itemize}


\section{Proposed Algorithm}
\label{sec:algo}
The effect of race has also been a widely studied topic in the field of machine-based face recognition~\cite{ge2009two, o1994structural, phillips2011other}. The existing literature has registered ethnicity as one of the important co-variates such as pose, illumination and expression. Notably, ethnicity co-exists with every other variate as it is a permanent attribute of a face unlike pose, illumination and expression which can be controlled. As for other face-related problems~\cite{dehon2001other, o1996other}, ethnicity of a face also becomes an element of consideration for facial retouching detection.
The demographics also affect the perception of beauty in humans~\cite{esther, rhodes2006evolutionary}. Due to this, the desired effects might be different in different demographics~\cite{zebrowitz1993they}, making it difficult for a detector built for subjects from one demographic to work with subjects from another. One method to make the system generalizable is to train the algorithm with `enough' data from every ethnicity~\cite{goldstein1985effects, lebrecht2009perceptual} but the data is difficult to obtain given the different regulatory constraints across countries. Another possibility is to enable the algorithm to learn these variations properly and utilize the knowledge to learn better representations. To achieve this for our algorithm, we utilize a classic idea from pattern recognition of utilizing class hierarchy to learn better data distributions~\cite{hastie1996discriminant, zhu2006subclass}.
 
The problem of retouching detection can be framed as a two-class classification problem where the two classes are original and retouched. In this paper, we extend the notion of the two classes (original and retouched) to be denoted by a hierarchy - classes (original and retouched) and subclasses (based on ethnicity and gender of the base face). The class hierarchy concept has been utilized in~\cite{hastie1996discriminant} to estimate a better (more discriminative) distribution of the underlying data using a mixture of Gaussians. This helps in capturing the variance of each of the sub-classes to further reduce the within class scatter and increase the between class variance. The work of Zhu and Martinez~\cite{zhu2006subclass} also adopts the same idea and allows the modeling of multiple distributions (depending on the number of subclasses) using the same formulation. It extends the prior work by proposing criteria to determine the number of subclasses in order to define the model.   

For our problem, the original and retouched are the two major classes while Caucasian, Chinese and Indian are the sub-classes within the major classes. The branching in class hierarchy can depend on the type of variations being captured by the data. The idea is to capture the correct distribution of the two classes based on demographics and use this estimate to obtain features that can better distinguish between the two major classes. To learn a discriminative representation between the two major classes that considers the variance of distribution between subclasses, we inherit the subclass distribution modeling into a sparse autoencoder, termed as Subclass Supervised Sparse Autoencoder ($S^3A$).



\begin{figure*}[t]
\centering
\includegraphics[scale=0.35]{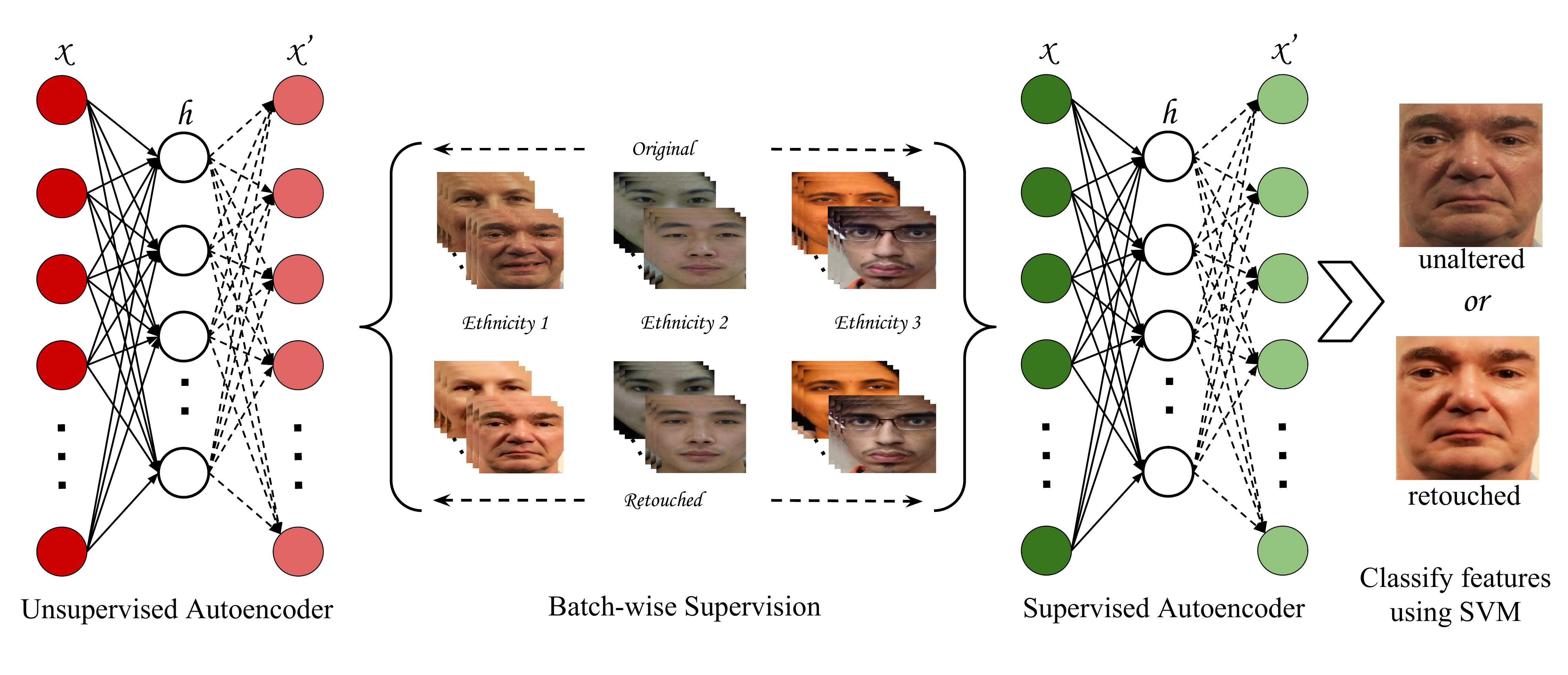}
   \caption{A diagram representing the steps involved in the proposed semi-supervised retouching detection algorithm.}
\label{fig:algo}
\end{figure*}

\subsection{Subclass Supervised Sparse Autoencoder}

An autoencoder is an unsupervised method to learn features that well represent the input data. For this purpose, the autoencoder has two parts -- the encoder and decoder. The encoder maps the input vector $x$ to a hidden representation $h=\phi(Wx)$ while the decoder maps the hidden representation to the reconstructed output $\hat{x} = W' \phi(Wx)$. Here, $W$ represents the weights between the input layer and the hidden layer and $\phi(.)$ is the activation function, usually a non-linear mapping function such as sigmoid. $W'$ are the weights between the hidden layer and reconstruction output layer. The weights $W$ and $W'$ are optimized based on the following cost function:




\begin{equation}
    \argminB_{W, W'} || X - W' \phi(WX) ||_2^2 + \lambda R
    \label{eqn:unsupervisedsparse}
\end{equation}

The cost function is the squared form of the reconstruction error (first term in equation \ref{eqn:unsupervisedsparse}) over the entire training set regularized with a function $R$. The training set is denoted by X where $X=\{x_1, x_2, x_3,...x_n\}$ and $x_i$ is an individual training instance. For the autoencoder to learn sparse representations, we can add the sparsity constraint through the regularization function. Regularization is the technique used to avoid overfitting in networks and make the networks more generalizable. It is based on adding terms to the objective function to penalize large weights. The term includes a regularization parameter $\lambda$ which regulates the amount of overfitting and a function $R(W, X)$ which together with $\lambda$ adjusts the cost function depending on $W$. The two popular regularization functions are the $l_2$ norm and the $l_1$ norm. Due to the nature of $l_1$ norm, most of the weights are suppressed to 0 and only a small subset of them remain non-zero. This is how the learned representations are sparse and robust to noise.


To improve discriminability in the learned features, Majumdar \etal proposed Class Sparsity based Supervised Encoding (CSSE)~\cite{majumdar2016face}. The aim of this method is to encode samples from the same class with same sparsity signature. This requires the class information of the training samples, thus making the autoencoder supervised. CSSE is achieved by organizing the training data as $X=\{x_{11},...x_{1n_1},x_{21},...x_{2n_2},...x_{c1},...x_{cn_c}\}$ where $x_{i1}...x_{in_i}$ are samples from the $i$th class and replacing the $l_1$ norm regularization function with a $l_{2,1}$ norm regularizer. The inner $l_2$ norm performs an aggregation within feature values while the outer $l_1$ norm encodes sparsity i.e., which features are enhanced for this particular set of samples. The main idea is to divide the data class-wise and then update the weights according to the sparsity constraint on each partition separately. This ensures the same level of sparsity in features belonging to the same class. The cost function of the CSSE is written as:

\begin{equation}
    \argminB_{W, W'} || X - W' \phi(WX) ||^2 + \lambda \sum_{c=1}^{C} || WX_c ||_{2,1}    
    \label{eqn:class}
\end{equation}

This method of class sparsity encoding treats all the samples of a class in a similar fashion. For problems that have a defined class hierarchy, we extend the idea in~\cite{majumdar2016face} to include subclass information while doing supervised encoding. This is achieved by regularizing the cost function based on the batches for each subclass. Hence, the updated cost function can be written as follows.



\begin{equation}
    \argminB_{W, W'} || X - W' \phi(WX) ||^2 + \lambda \Bigg[\sum_{i=1}^{c}\sum_{j=1}^{s_i} || WX_{ij} ||_{2,1}\Bigg]
    \label{eqn:subclass}
\end{equation}

Equation ~\ref{eqn:subclass} implies a two-level sparse encoding. The first level is for each subclass and the second level of sparsity imparts uniformity to the subclasses within a class. $l_{2,1}$ norm is non-differentiable but can be solved via approximation using Iterative Reweighted Least Squares (IRLS) approach. This uses the version of cost function in equation~\ref{eqn:subclasscont}. 

\begin{equation}
    \argminB_{W, W'} || X - W' \phi(WX) ||^2 + \lambda \Bigg[\sum_{i=1}^{c}\sum_{j=1}^{s_i} || \beta_{ij}WX_{ij} ||_2\Bigg]
    \label{eqn:subclasscont}
\end{equation}
Here, $\beta_{ij}$ is the set of parameters associated with each subclass. The weight update for each batch of data (samples within each subclass) is performed variably based on $\beta_{ij}$. The value of $\beta_{ij}$ is learned during training using M-Focuss algorithm~\cite{cotter2005sparse}. M-Focuss is a version of FOCal Underdetermined System Solver (Focuss) for multiple measurement vectors (MMV). It is a method to form sparse solutions to problems having solutions that share sparsity by reducing diversity. Thus, for a particular subclass, the optimization method will try to find solutions in the same neighborhood of parameter space (i.e. similar weights). 

\subsection{Retouching Detection Using \textbf{{$S^3A$}}}

For two-class classification problem of retouching detection \{\textit{original/unaltered}, \textit{retouched}\}, the cost function can be formulated as.

\begin{equation}
\begin{aligned}
    & \argminB_{W, W'} || X - W' \phi(WX) ||^2 \\
    & + \lambda \left[\sum_{i=1}^{s_1} || WX_i ||_{2,1} + \sum_{j=1}^{s_2} || WX_j ||_{2,1} \right]
\end{aligned}
\label{eqn:final}  
\end{equation}
where, first term is a standard autoencoder and the second term encodes the subclass variations belonging to two classes. Here $s_1$ and $s_2$ are the number of subclasses in two classes, respectively. 

As shown in Figure~\ref{fig:algo}, the proposed algorithm is trained  to determine if a given image is unaltered or retouched. For input of size $MN$, first, a two-hidden-layer network of size $[\frac{2}{3}MN, \frac{1}{2}MN]$ is pre-trained using normal face images from CMU Multi-PIE database~\cite{multipie} without using the subclass information. The network is fine-tuned using the training partition of MDRF database. For subclass based training, one experiment is performed with three ethnicity labels (Caucasian, Chinese, Indian) as three subclasses and another experiment is performed with gender (male, female) labels as two subclasses within each class. Note that, in this application, the subclasses are known, but if the subclasses are unknown then we can utilize subclass estimation techniques to label the training data with respect to subclasses. Once the model is trained, it is used to extract features for each image.  Using the features in the training set, a {2$\nu$-}SVM classifier~\cite{chew2005implementation} is trained to perform two-class classification with the original class being considered positive and the retouched class as negative.

\section{Experimental Results}
\label{sec:exp}
Face detection and normalization are performed on the data to be used in all the experiments. For the purpose of face detection, the detector from Viola and Jones~\cite{viola2004robust} is used. All the faces are tightly cropped and resized to $256\times256$. Once the images are pre-processed, they are used for analysis based on ethnicity of the trained retouching detection model and the analysis based on combined ethnicity trained models. Based on the different experimental protocols used, our experiments can be categorized into two parts.

\subsection{Experimental Protocols}

\begin{itemize}
\item{\textbf{Combined Evaluation}} -- To compare the performances of various retouching detection algorithms, the dataset is divided into two subject disjoint sets with equal number of male and female subjects belonging to each of the three ethnicities. Therefore, there are images from 300 subjects (150 males and 150 females) in each set. One of the sets is used for training and the other is used for testing. 

\item{\textbf{Cross Ethnicity Evaluation}} -- In order to analyze the differences of retouching among different ethnicities, we setup same-ethnicity training and testing, and other-ethnicity training and testing experiments. For training, 2 original images and 2 retouched images (one from Beauty Plus and one from Potrait Pro) are used for each subject in the set while all 6 images of remaining subjects are used for testing. The experiment uses 5-fold cross validation for same ethnicity and a $5\times5$-fold cross validation for different ethnicity. This implies that 5 trained models (trained with class balanced data) are obtained for each ethnicity using different combinations of the five folds. Each of the other remaining folds (full) for all three ethnicities are tested. In this way, we obtain 5 values of performance for same ethnicity and 25 for other ethnicities. An average of these performance values are reported along with standard deviation.

\end{itemize}

Comparison with state-of-the-art detection algorithms is performed on the MDRF database using the combined evaluation protocol. The specific algorithms used for comparison are explained below.
\vspace{10pt}

\noindent{\textbf{Makeup Detection}} -- In order to analyze the difference between the makeup and retouching, we compare the performance of an existing state-of-the-art makeup detection algorithm~\cite{kose2015facial} over the MDRF database. The algorithm uses Local Binary Pattern (LBP) and Histogram of Gradients (HOG) as features. The features upon normalization and concatenation are provided to two types of classifiers -- SVM (linear and with RBF kernel) and Alligator (a combination of classifiers) for classification. We use our implementation of the method based on the explanation provided by the authors and report classification accuracy of the method in detecting retouching. 

\noindent{\textbf{Photorealism Detection}} -- A huge amount of manipulation of the original photograph can make the photo look unrealistic which may be utilized to detect retouching.
One of the seminal works in detecting photorealism has been from Lyu and Farid~\cite{lyu2005realistic}. The authors use first-order and higher-order statistical characteristics of wavelets. Specifically, the wavelet decomposition of images is carried out using quadrature mirror filters. For each sub-band, at each scale and orientation, the mean, variance, skewness, and kurtosis are computed. These form the first half of the feature vector. The second half is comprised of statistics derived from the errors of linear predictor model that predicts magnitude of neighbouring coefficients. Once the features have been extracted from images, they are classified using SVM classifier. In this comparison, we consider the original class as photographic and the retouched class as photorealistic.

\noindent{\textbf{Spoofing Detection}} -- Morphing is a common method of spoofing. A recent proposed solution to accurately detect morphed images uses BSIF and a linear SVM~\cite{raghavendra2016detecting}. For the purposes of comparison, we use the same implementation~\cite{bsifcode} of BSIF as the authors and linear SVM from the libSVM package.

\noindent{\textbf{Facial Retouching Detection}} -- The detection algorithms mentioned so far use hand-crafted features. In order to compare with algorithms that use deep learned features, we evaluate the features from VGG-Face - a pre-trained convolutional neural network (CNN) model for faces~\cite{parkhi2015deep} with a SVM classifier. Another deep learning framework used for comparison is the facial retouching detection algorithm using Supervised Deep Boltzmann Machine (SDBM) and SVM by Bharati \etal~\cite{bharati2016detecting}. They use ~25,000 face images (original and retouched versions of images from Multi-PIE dataset) for training. For comparison purposes, the features are extracted using a trained SDBM model and a SVM is trained with features from the combined ethnicity training set of MDRF database.

\begin{figure}[t]
\centering
\hspace*{-0.2in}
\includegraphics[scale=0.42]{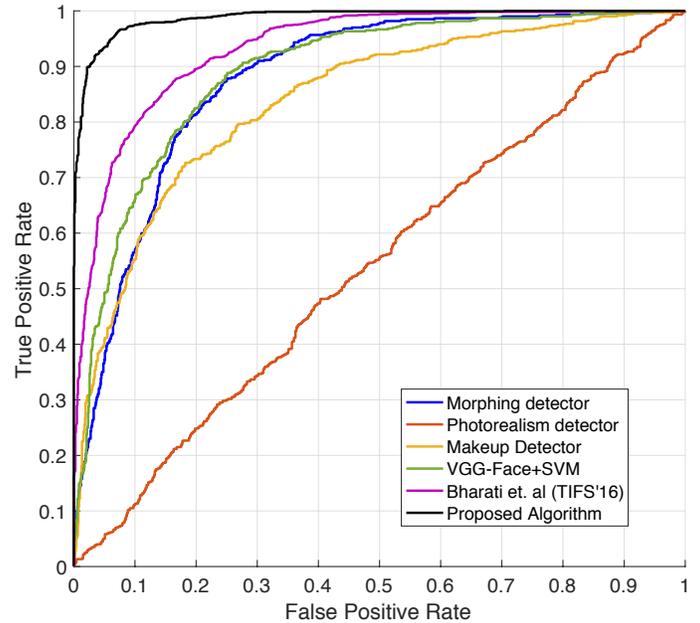}
\caption{Performance curves of detection algorithms with combined evaluation protocol on MDRF database.}
\label{fig:comparisonroc}
\end{figure}

\subsection{Observations for Combined Evaluation}
\label{subsec:results1}

The results of the proposed algorithm for retouching detection along with other detection algorithms have been demonstrated in Figure~\ref{fig:comparisonroc}. Some key observations based on the performance curves are:
\begin{itemize}
\item The specialized retouching detection algorithms perform better at retouching detection tasks than other detection algorithms. This shows that the other detection algorithms considered in this paper are not sufficient for generalized retouching detection.

\item The proposed algorithm outperforms the existing retouching detection algorithm by Bharati \etal. This demonstrates that incorporating subclass information can help estimate a better data distribution while learning representations. In this way, more discriminative (between classes) but generalizable (among subclasses) representations can be learned for each class using the proposed $S^3A$ algorithm. 


\item
In our results, descriptors learned from a supervised deep face recognition framework trained with VGG-Face are not as effective as the proposed algorithm in detecting retouching. As a part of the future work, this observation can be tested across other datasets and networks.

\end{itemize}

\begin{table}
\caption{Performance breakdown of retouching detection algorithms according to males and females and the two retouching tools.}
\vspace{-.5\baselineskip}

\begin{center}
\begin{tabular}{|l|c|c|c|c|c|}
\cline{2-5}
\multicolumn{1}{c|}{} & \multicolumn{2}{|c|}{Female} & \multicolumn{2}{|c|}{Male}  \\ \hline
Algorithm & Tool 1 & Tool 2 & Tool 1 & Tool 2 \\ \hline\hline
Bharati \etal~\cite{bharati2016detecting}& 84.1\% & 81.3\% & 90.3\% & 85.9\% \\
Proposed Algorithm & 94.2\% & 92.8\% & 95.9\% & 94.3\% \\
\hline
\end{tabular}
\end{center}
\label{tab:gendertool}
\end{table}

\begin{table*}
\caption{Results of cross ethnicity evaluation using VGG-Face+SVM, Bharati \etal~\cite{bharati2016detecting} and the proposed algorithm.}
\begin{center}
\begin{tabular}{|l|c|c|c|c|c|}
\cline{4-6}
\multicolumn{1}{c}{} & \multicolumn{1}{c}{} & \multicolumn{1}{c|}{} & \multicolumn{3}{|c|}{Testing Set}  \\ \cline{3-6}
\multicolumn{1}{c}{} & \multicolumn{1}{c|}{} & Algorithm & Caucasian & Chinese & Indian \\ \cline{4-6}\hline

\multirow{9}{3em}{Training Set} &  & VGG+SVM & 79.3\% ($\pm$ 3.1\%) & 81.1\% ($\pm$ 3.2\%) & 70.7\% ($\pm$ 2.5\%)\\
& Caucasian & Bharati \etal & 84.2\% ($\pm$ 2.9\%) &  80.9\% ($\pm$ 3.6\%) & 76.3\% ($\pm$ 4.1\%)\\
& & Proposed & 94.3\% ($\pm$ 1.1\%) & 91.9\% ($\pm$ 1.8\%) &  92.2\% ($\pm$ 2.1\%) \\ \cline{2-6}
& & VGG+SVM & 73.8\% ($\pm$ 2.5\%) & 87.0\% ($\pm$ 1.4\%) & 70.9\% ($\pm$ 4.6\%)\\
& Chinese & Bharati \etal & 79.4\% ($\pm$ 3.2\%) & 84.7\% ($\pm$ 2.2\%) & 80.1\% ($\pm$ 3.3\%)\\
& & Proposed  & 91.7\% ($\pm$ 2.3\%)& 97.5\% ($\pm$ 1.2\%) & 92.8\% ($\pm$ 2.2\%) \\ \cline{2-6}
& & VGG+SVM & 77.5\% ($\pm$ 2.8\%) & 81.4\% ($\pm$ 4.8\%) & 73.0\% ($\pm$ 4.1\%)\\
& Indian & Bharati \etal & 78.4\% ($\pm$ 3.7\%) & 79.3\% ($\pm$ 4.1\%) & 85.2\% ($\pm$ 2.9\%)\\
& & Proposed & 92.1\% ($\pm$ 2.3\%) & 93.6\% ($\pm$ 1.9\%) & 96.2\% ($\pm$ 1.4\%) \\
\hline
\end{tabular}
\end{center}
\label{tab:cross}
\end{table*}

To provide a more comprehensive understanding of the performance of the proposed algorithm with respect to existing state-of-the-art retouching detection, the breakdown of accuracies based on gender and tool has been provided in Table~\ref{tab:gendertool}. Tool 1 refers to BeautyPlus/MakeupPlus whereas Tool 2 refers to Potrait Pro Studio Max. The accuracy for male subjects is higher than the accuracy for female subjects. It is possible that, for females, the original face image might have pre-applied makeup which can lead to confusion between original and retouched class. According to the results, retouching for Tool 2 is slightly more difficult to detect than Tool 1.



\subsection{Observations for Cross-Ethnicity Evaluation}
\label{subsec:results2}

In order to analyze the performance for models trained on data from one ethnicity and tested on data from same ethnicity and different ethnicities, we performed experiments using the cross ethnicity evaluation protocol. Since we have 3 ethnicities in the MDRF database, we have 9 scenarios. The results for each of the scenarios are reported in Table~\ref{tab:cross}. The performance is reported for the 3 best performing algorithms in the combined evaluation. For each scenario and algorithm, the mean and standard deviation for classification accuracies of all trials have been reported.

According to the results of various scenarios, the following can be observed:

\begin{itemize}
\item As expected, the performance when the model is trained and tested on the same ethnicity (cells along the diagonal in table~\ref{tab:cross}) is generally higher than when the model is trained and tested on different ethnicities.
\item For all the scenarios, the proposed algorithm performs better than VGG+SVM and Bharati \etal~\cite{bharati2016detecting}. This shows that subclass sparsity encoding of gender helps in distinguishing between original and retouched images within a particular ethnicity. 
\item The standard deviation of accuracies across all folds and across all testing scenarios for each trained model is less for the proposed algorithm. This implies that the proposed technique generalizes better for multi-demographic retouching detection.
\item Of all the cross ethnicity evaluations (non-diagonal cells in Table~\ref{tab:cross}), retouching is easier to detect in Chinese faces while it is more difficult to detect in Indians.
\end{itemize}

The results presented in the paper show that the proposed algorithm improves upon the existing state-of-the-art by improving generalizability across demographics. The results also explain the relative difficulty in retouching detection among the three ethnicities - Caucasian, Chinese and Indian. 

\section{Conclusion \& Future Work}
The paper presents the limitations of state-of-the-art algorithms in detecting retouching for multiple demographics. To achieve this, a novel dataset with subjects from three ethnicities and retouching from two tools is introduced. Combined ethnicity and cross-ethnicity evaluations were performed to obtain a detailed understanding of performance in scenarios with one or more ethnicities available for training. Finally, a novel semi-supervised framework with Subclass Supervised Sparse Autoencoder ($S^3A$) is proposed to improve detection of retouching across ethnicity and gender. 

Some possible avenues of future research include analyzing differences among tools from various demographic area and to create generalized algorithms. Also, quantifying how much retouching is acceptable and beyond what levels is retouching considered spoofing can be beneficial for ensuring standards for images.

{\small
\bibliographystyle{ieee}
\bibliography{bibliography}
}

\end{document}